\documentclass[conference]{IEEEtran}
\IEEEoverridecommandlockouts

\usepackage{cite}
\usepackage{amsmath,amssymb,amsfonts}
\usepackage{graphicx}
\usepackage{textcomp}
\usepackage{xcolor}

\usepackage[T1]{fontenc}
\usepackage{subfig}
\usepackage{algorithm}
\usepackage{hyperref}
\usepackage{adjustbox}
\usepackage{cuted}
\usepackage{mathtools}
\usepackage{makecell}
\newtheorem{definition}{Definition}
\usepackage{algpseudocodex}

\def\BibTeX{{\rm B\kern-.05em{\sc i\kern-.025em b}\kern-.08em
    T\kern-.1667em\lower.7ex\hbox{E}\kern-.125emX}}

\newenvironment{Acknowledgments}
    {\section*{Acknowledgments} }
    {}
    
\begin{document}
\title{Joint Explainability-Performance Optimization with Surrogate Models
for AI-Driven Edge Services}

\author{\IEEEauthorblockN{\textit{Foivos Charalampakos$^*$,
Thomas Tsouparopoulos$^*$,
Iordanis Koutsopoulos}}
\IEEEauthorblockA{\textit{Department of Informatics,}
\textit{Athens University of Economics and Business}\\
\textit{Athens, Greece} \\
}
}

\maketitle

\begingroup\renewcommand\thefootnote{*}
\footnotetext{Equal contribution}
\endgroup

\begin{abstract}
Explainable AI is a crucial component for edge services, as it ensures reliable decision making based on complex AI models. \textit{Surrogate models} are a prominent approach of XAI where human-interpretable models, such as a linear regression model,  are trained to approximate a complex (black-box) model's predictions. This paper delves into the balance between the predictive accuracy of complex AI models and their approximation by surrogate ones, advocating that both these models benefit from being learned simultaneously. We derive a joint (bi-level) training scheme for both models and we introduce a new algorithm based on multi-objective optimization (MOO) to simultaneously minimize both the complex model’s prediction error and the error between its outputs and those of the surrogate. Our approach leads to improvements that exceed $\mathbf{99\%}$ in the approximation of the black-box model through the surrogate one, as measured by the metric of \textit{Fidelity}, for a compromise of less than $\mathbf{3\%}$ absolute reduction in the black-box model’s predictive accuracy, compared to single-task and multi-task learning baselines. By improving \textit{Fidelity}, we can derive more trustworthy explanations of the complex model’s outcomes from the surrogate, enabling reliable AI applications for intelligent services at the network edge.

\end{abstract}

\section{Introduction}

Deploying intelligent services at the network edge, e.g., at home gateways, micro-servers, or small cells, is crucial for reducing latency and improving Quality of Service (QoS) for users. These services leverage data generated by end devices to train sophisticated machine learning (ML) models that enhance user experiences in applications such as mobile Augmented/Virtual Reality, cognitive voice or text-based personal assistants powered by Large Language Models, and mobile health monitoring systems. However, while complex ML models, such as Neural Networks (NNs), excel in making predictions with high accuracy they lack transparency, which limits their applicability.

In this evolving landscape, explainability must be considered alongside accuracy when designing ML models for intelligent edge services. Beyond achieving high predictive performance, ML models should provide insights into their decision-making process to ensure trust and accountability. For example, in the case of a cognitive personal assistant, explanation for a specific AI-generated advice can be provided to the user. The field of EXplainable Artificial Intelligence (XAI) aims to elucidate the decision-making processes of complex ML models and address these challenges.

A prominent class of XAI methods is that of \textit{surrogate models} which approximate complex, opaque models through simpler, interpretable ones \cite{Ribeiro}. Surrogate models serve as interpretable proxies, allowing operators to understand how input variables influence the ML model’s predictions. A surrogate model can be any inherently interpretable model such as a linear function that offers feature importance scores via the learned coefficients, or a decision tree that simulates its prediction through a corresponding decision rule. In this work, \textit{our goal is to train a ML model that can be accurately approximated by an interpretable surrogate, without significantly compromising its predictive performance}. This balance is crucial for intelligent services at the edge-cloud continuum, where insights from ML models drive informed actions.

In the typical setting, surrogate models are trained on the black-box model’s predictions \textit{after} training of the black-box model and thus, the two distinct objectives of approximation accuracy for the surrogate model and predictive accuracy for the complex one are optimized independently. In recent works, e.g., \cite{Charalampakos}, Multi-Task Learning (MTL) was employed with the aim of learning simultaneously these two objectives by combining them into a joint loss function (a weighted linear sum). This approach improved the surrogate model’s approximation at a high cost of the complex model's accuracy due to the trade-off between these two conflicting objectives. 

Achieving both prediction accuracy and explainability at the same time necessitates the adoption of a Multi-objective optimization (MOO) framework. MOO addresses problems with conflicting objectives, such as those in MTL, more effectively than simply combining objectives in one, as it captures the full interaction and trade-offs. MOO methods identify a \textit{Pareto front} \cite{Censor}, namely a set of solutions representing different compromises between task-specific objectives, where improvements in one objective come at the expense of lowering the performance of another objective. This is directly relevant to the trade-off between approximation accuracy and predictive accuracy that we aim to improve by jointly training black-box and surrogate models.

This paper offers contributions to the field of XAI for edge services by framing the trade-off between the objectives of approximation accuracy of the surrogate model and predictive accuracy of the complex ML model as a MOO problem. We shift the focus from typical MOO-based approaches \cite{Désidéri, Sener} that learn conflicting tasks to a two-level nested MOO framework where the surrogate model and the black-box model are optimized simultaneously. This nested formulation is more intricate since the surrogate model's approximation directly influences the training of the black-box model, creating a feedback loop that forces the latter to become more ``explainable'' without substantial loss in accuracy.

The quality of approximation of the black-box model by the surrogate one is captured by Fidelity, a metric that indirectly caters for explainability by measuring the disagreement between the predictions of the two models (thus, lower is better) \cite{Plumb}. \\
The contributions of this work can be summarized as follows: 
\begin{itemize}
\item We introduce a novel joint training framework that simultaneously optimizes both a black-box model and its interpretable surrogate, by formulating the problem as a MOO problem.
\item We perform an ablation study which shows that decoupling objectives harms approximation, emphasizing the need for joint MOO-based training.
\item We demonstrate that our approach outperforms single-task and multi-task baselines across various datasets in both global and local explainability settings, with notable Fidelity improvements for more reliable explanations and minimal predictive accuracy loss.
\end{itemize}

\section{Related Work}

\label{Rel:work}

\textbf{Surrogate models in XAI:} Surrogate models have been used in the literature as a means to approximate the predictions of black-box AI models, and they are categorized into global and local surrogates. Global surrogate models \cite{Charalampakos} aim at approximating the behavior of the black-box in the entire dataset, while local surrogate models \cite{Ribeiro} are trained to provide explanations for individual instances of the dataset. 

Our work focuses primarily on global surrogate models, which are typically trained using the predictions of a given black-box model over a training dataset, but also extends to the local explainability setting. This is achieved by minimizing a loss function, known in the literature as \textit{Point Fidelity} \cite{Plumb}, that measures the (dis)agreement between the output of the black-box model and the output of the surrogate one. \\
\textbf{Multi-Objective Optimization (MOO):} MOO addresses the problem of optimizing a set of possibly contrasting objectives in order to find a set of solutions where in each solution, no objective can be improved without degrading at least one other objective \cite{Kuhn}. In \cite{Schneider}, the optimization of accuracy and interpretability-based metrics is addressed with MOO, however its gradient-free algorithm scales poorly with the number of parameters found in most modern NNs. Another line of work \cite{Navon}, utilizes hyper-networks to approximate the entire Pareto front given a preference weight vector as input. 

\par Gradient-based MOO algorithms compute gradients of multiple loss functions to find Pareto optimal points that balance the trade-off between objectives. Based on the Karush-Kuhn-Tucker (KKT) conditions \cite{Kuhn}, they seek a descent direction for the gradients that optimizes all objectives simultaneously. In \cite{Désidéri}, the authors proposed the Multiple Gradient Descent Algorithm (MGDA) which converges to a local Pareto optimal point by using a descent direction from the convex hull of the tasks' gradients. In \cite{Sener}, the authors formulate MTL as a MOO problem and modify MGDA to scale for large-scale NNs. On a similar note, the approach in \cite{Liu} manipulates the update direction to find a better optimization trajectory and converges to the optimal point. 
\par Our work diversifies from empirical MTL-based solutions by employing  MOO to jointly train a complex model and an interpretable, surrogate one. We formulate a bi-level optimization algorithm to cater for the concurrent optimization of the two objectives, namely that of predictive accuracy, and the approximation of the complex model's output via the surrogate model. We apply gradient-based optimization to find local Pareto optimal solutions, since it scales efficiently for NNs. Additionally, the black-box model which is optimized by our algorithm can be used to obtain local surrogate models that more accurately approximate its decision boundary for specific data instances, compared to black-box models optimized with conventional approaches.

\section{Pareto Optimality in MOO: A Primer}
We define a typical MOO problem which consists of $T$ objectives captured by the respective loss functions $\mathcal{L}_t:\mathbb{R}^{d}\to\mathbb{R}^{+}, \,\, t = 1,...,T$, ($\mathcal{L}:\mathbb{R}^{d}\to\mathbb{R}^{T}$ in vector form), where $d$ is the dimension of the input vector. The general objective of (parametric) MOO takes the following form:
\begin{equation}
\min_{\boldsymbol{\theta} \in \mathbb{R}^{m}} \mathcal{L}(\boldsymbol{\theta}) = (\mathcal{L}_1(\boldsymbol{\theta}), \mathcal{L}_2(\boldsymbol{\theta}), ..., \mathcal{L}_T(\boldsymbol{\theta})) \,.
\end{equation}
The goal of MOO is to optimize all objectives simultaneously, i.e., find an optimal solution $\boldsymbol{\theta}$ that minimizes all the objective functions $L_t(\cdot)$ for $t = 1,...,T$. The following definitions are central to MOO.
\begin{definition}
\textbf{Pareto dominance: }Let $\boldsymbol{\theta}_1, \boldsymbol{\theta}_2$ be two feasible solutions in $\mathbb{R}^{m}$, then we say that $\boldsymbol{\theta}_1$ dominates $\boldsymbol{\theta}_2$ ($\boldsymbol{\theta}_1 \prec \boldsymbol{\theta}_2$) if and only if  $\forall\, i \in T: \,\,\mathcal{L}_i(\boldsymbol{\theta}_1) \leq \mathcal{L}_i(\boldsymbol{\theta}_2)$ and $\exists\, j\in T: \mathcal{L}_j(\boldsymbol{\theta}_1) < \mathcal{L}_j(\boldsymbol{\theta}_2)$.
\end{definition}

Intuitively, if a point $\boldsymbol{\theta}_1$ dominates $\boldsymbol{\theta}_2$, then $\boldsymbol{\theta}_1$ is clearly preferable, because it improves at least one objective without worsening any other objective. 
\begin{definition}
\textbf{Pareto optimality: }A solution $\boldsymbol{\theta}^{*} \in \mathbb{R}^{m}$ is a Pareto optimal point if $\nexists\,\hat{\boldsymbol{\theta}} \in \mathbb{R}^{m}$ such that $\hat{\boldsymbol{\theta}} \prec \boldsymbol{\theta}^{*}$.
\end{definition}
\begin{definition}
\textbf{Pareto front: }The set of all Pareto optimal points is called the Pareto front.
\end{definition}

When the objective functions are convex, gradient-based algorithms for MOO converge to globally optimal Pareto solutions. However, in many modern ML applications, achieving optimal performance often relies on deep NNs, thus leading to non-convex training loss functions with respect to the NN's weights.  This results in a non-convex optimization problem where attaining global optimality is not feasible. We thus give the following definition of \textit{local} Pareto optimality: 
\begin{definition}
\textbf{Local Pareto optimality: } A solution $\boldsymbol{\theta}^{*} \in \mathbb{R}^{m}$ is called \textit{local Pareto optimal} if there exists a neighborhood of $\,\, \boldsymbol{\theta}^{*},\,\, \mathcal{N(\boldsymbol{\theta}^{*})} \subset \mathbb{R}^{m}: \forall \,\, \boldsymbol{\theta} \in \mathcal{N(\boldsymbol{\theta}^{*})} \setminus \boldsymbol{\theta}^{*}, \boldsymbol{\theta} \nprec \boldsymbol{\theta}^{*}$, i.e., no other solution in its neighborhood can have better values in all objective functions.
\end{definition}

\section{System Model and Proposed Framework}
\label{Problem}
In this section, we introduce our MOO problem formulation and outline the accompanying optimization algorithm.
\subsection{System Model}
\label{Background}

In the general setting, datasets will reside in different locations and will need to be fused in other locations in order to give rise to a trained ML model. In this work, we set off by considering the simplest instance of the problem, where the dataset of interest resides in one location e.g., a gateway, and an ML model needs to be trained out of this data, albeit with the MOO perspective of prediction accuracy and explainability. We jointly train a black-box model with the purpose of solving a ML task, e.g., classification or regression and a surrogate model, which has an inherently explainable structure, to approximate the black-box model's predictions. 
\par Concretely, let $f_{\boldsymbol{\theta}}: \mathbb{R}^{d}\to\mathbb{R}^{+}$ be a black-box model, parameterized by $\boldsymbol{\theta}$, we aim to explain through surrogate approximation. Let $\mathcal{G} = \{ g: \mathbb{R}^{d}\to\mathbb{R}^{+} : g \,\, \text{is differentiable}\}$ be the class of \textit{differentiable, inherently interpretable} functions used to approximate the prediction of $f_{\boldsymbol{\theta}}(\cdot)$. Note that the constraint of the surrogate's differentiability is essential for the gradient-based training, and we denote these functions parameterized by $\boldsymbol{\phi}$ as $g_{\boldsymbol{\phi}}(\cdot)$ hereafter. For example, given an input vector \( \boldsymbol{x} \in \mathbb{R}^d \), $g_{\boldsymbol{\phi}}(\cdot)$ can be a linear function: $g_{\boldsymbol{\phi}}(\boldsymbol{x}) = \mathbf{\boldsymbol{\phi}}^\top \boldsymbol{x} + b$ or a differentiable decision tree \cite{Suarez} which uses a logistic function $\sigma(\mathbf{\boldsymbol{\phi}}^\top \boldsymbol{x} + b)$ to split the feature space into binary tree-like branches.

We aim to balance the predictive accuracy of the complex model $f_{\boldsymbol{\theta}}(\cdot)$ and the accuracy of its approximation via the surrogate model $g_{\boldsymbol{\phi}}(\cdot)$
by formulating a bi-level multi-objective optimization objective.
\sloppy The respective losses for the two models can be denoted as $\mathcal{L}_{\text{\tiny pred}}(\boldsymbol{\theta})$ and $\mathcal{L}_{\text{\tiny PF}}(\boldsymbol{\theta}, \boldsymbol{\phi})$, where $\mathcal{L}_{\text{\tiny pred}}(\boldsymbol{\theta}) = \mathcal{L}_{\text{\tiny pred}} (f_{\boldsymbol{\theta}}(\boldsymbol{x}), y)$ is the loss function for the predictive task (e.g., Cross-entropy for classification tasks) and $\mathcal{L}_{\text{\tiny PF}}(\boldsymbol{\theta}, \boldsymbol{\phi}) = \mathcal{L}_{\text{\tiny PF}}(f_{\boldsymbol{\theta}}(\boldsymbol{x}), g_{\boldsymbol{\phi}}(\boldsymbol{x})) = (f_{\boldsymbol{\theta}}(\boldsymbol{x}) - g_{\boldsymbol{\phi}}(\boldsymbol{x})) ^ 2$ is \textit{Point Fidelity} \cite{Plumb}, which measures how close the predictions of $g_{\boldsymbol{\phi}}(\cdot)$ are to those of $f_{\boldsymbol{\theta}}(\cdot)$, i.e., difference between the output of the two models. Our goal is to minimize the training loss of $f_{\boldsymbol{\theta}}(\cdot)$, while ensuring that it can be well approximated by the surrogate model $g_{\boldsymbol{\phi}}(\cdot)$.
This translates into the following bi-level MOO problem involving a MOO problem for the parameters of the black-box model at the upper-level and the minimization of the loss function of the surrogate model at the lower-level:
\begin{equation}
    \label{eq:ΜΑΙΝ_ΜΟΟ}
    \begin{aligned}
    \min_{\boldsymbol{\theta}} \left[\mathcal{L}_{\text{\tiny pred}} (\boldsymbol{\theta}), \mathcal{L}_{\text{\tiny PF}}(\boldsymbol{\theta}, \boldsymbol{\phi^*}) \right] \\
    \text{s.t. } \boldsymbol{\phi^*} \in \mathop{\mathrm{arg\,min}}_{\boldsymbol{\phi}} \mathcal{L}_{\text{\tiny PF}}(\boldsymbol{\theta}, \boldsymbol{\phi})\,.
    \end{aligned}
\end{equation}

This problem formulation encapsulates the predictive accuracy - approximability trade-off by optimizing $\boldsymbol{\theta}$ to concurrently decrease the training predictive loss and the disagreement in the predictions between $f_{\boldsymbol{\theta}}(\cdot)$ and $g_{\boldsymbol{\phi}}(\cdot)$, while at the same time finding the optimal set of parameters $\boldsymbol{\phi}$ for the surrogate model. Thus, we have to solve a MOO problem for $\boldsymbol{\theta}$, subject to optimizing $\boldsymbol{\phi}$. This problem can be solved with a technique such as stochastic gradient descent ({SGD}).

\begin{figure}[t!]
\centering
\includegraphics[scale=0.3]{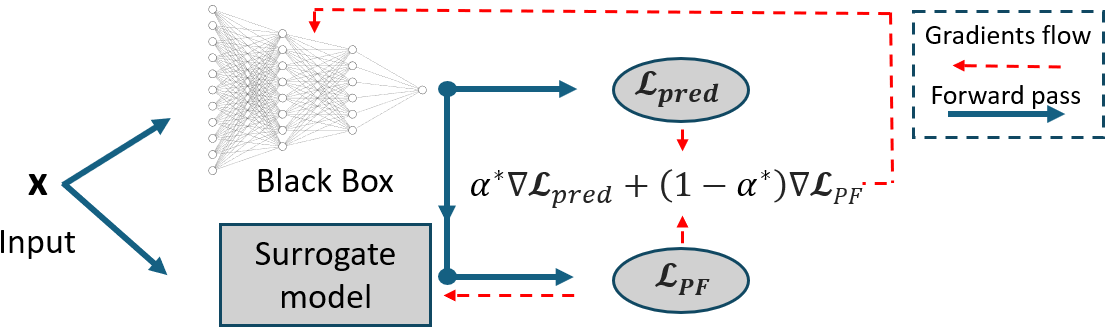}
\caption{The proposed MOO framework. Input data $\boldsymbol{x}$ is passed through the black-box and the surrogate model. The outputs of the former are used to calculate the two losses, $\mathcal{L}_{\text{\tiny pred}}$ and $\mathcal{L}_{\text{\tiny PF}}$. Then, the red dashed lines show  the gradients flow that result from our gradient-based optimization.}\label{fig:framework}
\end{figure}

\subsection{Our Proposed Framework}
To solve the MOO problem with respect to parameters $\boldsymbol{\theta}$, we apply the \textit{MGDA} algorithm which was introduced in \cite{Désidéri}.
At each iteration, \textit{MGDA} finds the direction of the gradient that improves both objectives of the upper-level in~(\ref{eq:ΜΑΙΝ_ΜΟΟ}). For two objectives, the following convex quadratic problem is optimized:
\begin{equation}
\label{eq:alpha_min}
\min_{\alpha\in(0,1)}||\alpha \cdot \nabla_{\boldsymbol{\theta}} \mathcal{L}_{\text{\tiny pred}} (\boldsymbol{\theta}) + (1 - \alpha) \cdot \nabla_{\boldsymbol{\theta}} \mathcal{L}_{\text{\tiny PF}}(\boldsymbol{\theta}, \boldsymbol{\phi})||^2_2 \,.
\end{equation}
Hence, we solve~(\ref{eq:alpha_min}) to obtain a solution $\alpha^*$ which is used as the scaling factor for the gradients of the two losses to update the parameters $\boldsymbol{\theta}$ of the black-box model. In our setting with two objectives, $\alpha^*$ can be obtained as the closed-form solution of the convex problem (\ref{eq:alpha_min}):

\begin{equation}
    \label{eq:alpha}
    \alpha^* = \left[ \frac{\left( \nabla_{\boldsymbol{\theta}} \mathcal{L}_{\text{\tiny PF}}(\boldsymbol{\theta}, \boldsymbol{\phi}) - \nabla_{\boldsymbol{\theta}} \mathcal{L}_{pred} (\boldsymbol{\theta})^\intercal \nabla_{\boldsymbol{\theta}} \mathcal{L}_{\text{\tiny PF}}(\boldsymbol{\theta}, \boldsymbol{\phi})\right) }{||\nabla_{\boldsymbol{\theta}} \mathcal{L}_{pred} (\boldsymbol{\theta}) - \nabla_{\boldsymbol{\theta}} \mathcal{L}_{\text{\tiny PF}}(\boldsymbol{\theta}, \boldsymbol{\phi})||_2^2} \right]_{+
}
\end{equation} 
where $[\cdot]_{+, 
} = \max(\min(\cdot, 1),
0)$ represents the clipping operation in $[0, 1]$. In \cite{Désidéri}, the authors showed that the solution to~(\ref{eq:alpha_min}) is either $0$, leading to a \textit{Pareto stationary} point (i.e., {the gradients with respect to all objectives are equal to $0$}), or it offers a descent direction that enhances performance across all tasks. A schematic depiction of the proposed framework is shown in Fig. \ref{fig:framework}. The black-box model and the surrogate one are trained concurrently, with the black-box model's parameters updated using gradients scaled by the solution of~(\ref{eq:alpha_min}).

\begin{algorithm}[t!]
\caption{Our MOO-based approach}\label{Alg:Alg1}
\begin{algorithmic}
\State \textbf{Input:} training set $\{\boldsymbol{x}_i, y_i\}_{i=1}^N$,  black-box model $\boldsymbol{\theta}$, surrogate model $\boldsymbol{\phi}$, learning rates $\eta_{\boldsymbol{\theta}}$ and $\eta_{\boldsymbol{\phi}}$ 
\State \textbf{Output:} Optimized model parameters $\boldsymbol{\theta}^*$ and $\boldsymbol{\phi}^*$
\While{$\boldsymbol{\theta}$ not in a local Pareto optimal point}
\State $\mathbf{x}, \mathbf{y} = \{(\boldsymbol{x}_i, y_i)\}_{i=1}^{N}$
        \State Calculate $\mathcal{L}_{\text{\tiny PF}}(f_{\boldsymbol{\theta}}(\mathbf{x}), g_{\boldsymbol{\phi}}(\mathbf{x}))$ using fixed $\boldsymbol{\theta}$
        \State$\boldsymbol{\phi} \gets \boldsymbol{\phi} - \eta_{\boldsymbol{\phi}} \cdot \nabla_{\boldsymbol{\phi}} \mathcal{L}_{\text{\tiny PF}}$     \Comment{Outputs $\boldsymbol{\boldsymbol{\phi}}^*$}
    \State{Solve~(\ref{eq:alpha_min}) to obtain $\alpha^*$}

    \State{Calculate $\boldsymbol{d_{\boldsymbol{\theta}}} = \alpha^* \cdot \nabla_{\boldsymbol{\theta}} \mathcal{L}_{\text{\tiny pred}} (f_{\boldsymbol{\theta}}(\mathbf{x}), \mathbf{y})\,\, + \,\,(1 - \alpha^*) \cdot \nabla_{\boldsymbol{\theta}} \mathcal{L}_{\text{\tiny PF}}(f_{\boldsymbol{\theta}}(\mathbf{x}), g_{\boldsymbol{\phi}}(\mathbf{x}))$ }
   
    \State{$\boldsymbol{\theta} \gets \boldsymbol{\theta} - \eta_{\boldsymbol{\theta}}\cdot \boldsymbol{d_{\boldsymbol{\theta}}}$}  
    \Comment{Outputs $\boldsymbol{\theta}^*$}
\EndWhile
\end{algorithmic}
\end{algorithm}

We present an iterative, gradient-based optimization algorithm for our MOO formulation in Algorithm~\ref{Alg:Alg1}, to optimize the parameters of models $f_{\boldsymbol{\theta}}(\cdot)$ and $g_{\boldsymbol{\phi}}(\cdot)$. The algorithm updates $\boldsymbol{\theta}$ until convergence to the local Pareto optimal point. Within each iteration, $\boldsymbol{\phi}$ is also updated (with $\boldsymbol{\theta}$ fixed), with the goal of minimizing $\mathcal{L}_{\text{\tiny PF}}$. Then, the algorithm computes $\alpha^*$ according to~(\ref{eq:alpha}) and derives the gradient $\boldsymbol{d}_{\boldsymbol{\theta}}$, as a weighted combination of gradients (w.r.t. $\boldsymbol{\theta}$) from both the predictive loss function $\mathcal{L}_{\text{\tiny pred}}(\cdot, \cdot)$ and the fidelity-based loss function $\mathcal{L}_{\text{\tiny PF}}(\cdot, \cdot)$. The parameters of the black-box model, $\boldsymbol{\theta}$ are then updated using gradient descent across the direction of $\boldsymbol{d}_{\boldsymbol{\theta}}$. This process continues until convergence, and the resulting parameter sets, $\boldsymbol{\phi}^*$ and $\boldsymbol{\theta}^*$ are acquired.

\section{Experiments}
\label{Experiments}
In this section, we present the evaluation results of our algorithm in classification and regression tasks. 

\subsection{Experimental Setup}

\textbf{Models}. For the black-box model $f_{\boldsymbol{\theta}}$, we focus on Multi-Layer Perceptrons (MLPs), where the number of hidden layers and neurons are chosen based on performance on the validation set of each dataset. 
Potentially, any differentiable architecture could serve as a candidate for $f_{\boldsymbol{\theta}}$. For the surrogate model $g_{\boldsymbol{\phi}}$, we opt for a linear function of the input features. Concretely, given an input vector \( \boldsymbol{x} \in \mathbb{R}^d \), a linear surrogate model takes the form: \( g_{\boldsymbol{\phi}}(\mathbf{x}) = \mathbf{\boldsymbol{\phi}}^\top \boldsymbol{x} + b \), where \( \boldsymbol{\phi} \in \mathbb{R}^d \) represents the coefficients (weights), and \( b \in \mathbb{R} \) is the bias term. Although we use a linear model, any other differentiable model can be used, provided it is explainable. For the optimization, Adam \cite{Adam} is employed with learning rates $\eta_{\boldsymbol{\theta}} = \eta_{\boldsymbol{\phi}} = 10^{-3}$. The hyper-parameter values were chosen based on a hyper-parameter grid search on the validation set. \\
\textbf{Datasets}. We test our models on regression and classification tasks, with the UCI database\footnote{\href{https://archive.ics.uci.edu}{https://archive.ics.uci.edu}}, and the California Housing dataset\cite{Pace}. For each dataset, we standardize numerical features to have mean zero and variance one and one-hot encode the categorical ones. For imaging data, we use the MNIST dataset \cite{Deng}, and standardize the pixels to take values in $[0, 1]$. \\
\textbf{Baselines}. 
The baselines we consider for our experiments are: 
\begin{itemize}
\item \textit{linear:} a linear model for predictions,
\item \textit{single-task learning (STL):} a black-box model $f_{\boldsymbol{\theta}}$ is trained and then, a global surrogate model $g_{\boldsymbol{\phi}}$ is trained to approximate its predictions,
\item  \textit{uniform scaling (UNI):} minimize a uniformly weighted sum of the two losses, i.e., $\frac{1}{2} \mathcal{L}_{\text{\tiny pred}} (\cdot, \cdot) + \frac{1}{2}\mathcal{L}_{\text{\tiny PF}}(\cdot, \cdot)$, 
\item  \textit{grid search (GS):} grid search for various values of $\alpha \in (0, 1)$ to weight the common loss $\alpha\cdot \mathcal{L}_{\text{\tiny pred}}(\cdot, \cdot) + (1 - \alpha)\cdot\mathcal{L}_{\text{\tiny PF}}(\cdot, \cdot)$. The values are fixed once, before the initialization of the optimization process (based on \cite{Charalampakos}), 
\item  \textit{random step search (RND):} a different random value of $\alpha \in (0, 1)$ is uniformly sampled at each step of the optimization process to weigh the common loss $\alpha \cdot\mathcal{L}_{\text{\tiny pred}}(\cdot, \cdot) + (1 - \alpha)\cdot\mathcal{L}_{\text{\tiny PF}}(\cdot, \cdot)$. 
\end{itemize} 
\begin{table*}[t!]
\centering
\caption{\label{table:baselines} Comparison of our approach with {\textit{STL}} and \textit{MTL} (\textit{UNI}, \textit{RND}) baselines, using the task-specific metric for predictive performance and $\mathrm{GF}$. The results for \textit{Linear} are empty in $\mathrm{GF}$ because Fidelity equals to zero.}

\begin{adjustbox}{width=\textwidth}
\begin{tabular}{|l|ccccc|c}
\cline{1-6}
\multicolumn{1}{|c|}{\bf Dataset} &
  \multicolumn{1}{c|}{\bf Linear} &
  \multicolumn{1}{c|}{\bf STL} &
  \multicolumn{3}{c|}{\bf MTL} &
   \\ \cline{1-6}
  \multicolumn{1}{|c|}{} &
  \multicolumn{1}{c|}{} &
  \multicolumn{1}{c|}{} &
  \multicolumn{1}{c|}{\bf UNI } &
  \multicolumn{1}{c|}{\bf RND} &
  \multicolumn{1}{c|}{\bf Ours} 
   \\ \cline{1-6}
\multicolumn{1}{|c|}
{\begin{tabular}[c]{@{}l@{}} \sc power (mse)  \\ \sc adult ($F_1$)\\ \sc housing (mse)\\ \sc magic ($F_1$)\\\end{tabular}} & 
\multicolumn{1}{c|}{\begin{tabular}[c]{@{}c@{}} $0.08$ ($\pm0.001$) \\ $0.72$ ($\pm0.002$) \\ $0.43$ ($\pm0.009$) \\ $0.78$ ($\pm0.001$) \end{tabular}} &
\multicolumn{1}{c|}{\begin{tabular}[c]{@{}c@{}}$0.05$ ($\pm0.004$) \\ $0.78$ ($\pm0.02$) \\$0.23$ ($\pm0.02$)\\ $0.86$ ($\pm0.002$)\end{tabular}} &
\multicolumn{1}{c|}{\begin{tabular}[c]{@{}c@{}}$0.06 $ ($\pm0.002$)\\ $0.77$ ($\pm0.007$) \\ $0.30$ ($\pm0.004$) \\ $0.80$ ($\pm0.004$)\end{tabular}} &
\multicolumn{1}{c|}{\begin{tabular}[c]{@{}c@{}} $0.06$ ($\pm0.003$)\\ $0.78$ ($\pm0.003$) \\ $0.30$ ($\pm0.005$)\\ $ 0.80$ ($\pm0.004$)\end{tabular}} &
\multicolumn{1}{c|}{\begin{tabular}[c]{@{}c@{}}$0.06$ ($\pm0.002$)\\ $0.77$ ($\pm0.001$)\\ $0.27$ ($\pm0.003$) \\ $ 0.82$ ($\pm0.001$)\end{tabular}} &
\\ \cline{1-6}
\multicolumn{1}{|c|} 
{\begin{tabular}[c]{@{}l@{}} \sc power ($\mathrm{GF}$) \\ \sc adult ($\mathrm{GF}$)\\ \sc housing ($\mathrm{GF}$) \\ \sc magic ($\mathrm{GF}$) \\\end{tabular}} & 
\multicolumn{1}{c|}{\begin{tabular}[c]{@{}c@{}} - \\ - \\ - \\ - \end{tabular}} &
\multicolumn{1}{c|}{\begin{tabular}[c]{@{}c@{}}$0.02$ ($\pm0.003$)\\ $0.08$ ($\pm0.02$) \\$0.31$ ($\pm0.05$)\\ $0.11$ ($\pm0.001$)\end{tabular}} &
\multicolumn{1}{c|}{\begin{tabular}[c]{@{}c@{}}$0.015$ ($\pm0.004$) \\$ 0.01$ ($\pm0.008$) \\ $0.03$ ($\pm0.02$)\\ $0.04$ ($\pm0.001$)\end{tabular}} &
\multicolumn{1}{c|}{\begin{tabular}[c]{@{}c@{}} $0.003$ ($\pm0.05$)\\ $0.02$ ($\pm0.008$) \\ $0.02$ ($\pm0.01$) \\ $0.06$ ($\pm0.002$)\end{tabular}} &
\multicolumn{1}{c|}{\begin{tabular}[c]{@{}c@{}} $\bf8.91\cdot10^{-4}$ ($\bf\pm0.0015$) \\  $\bf1.2\cdot10^{-5}$ ($\bf\pm0.004$) \\ $\bf0.014$ ($\bf\pm0.02$) \\ $\bf3\cdot10^{-5}$ ($\bf\pm0.001$)\end{tabular}} &    \\ \cline{1-6}
\end{tabular}
\end{adjustbox}
\end{table*}
\textbf{Evaluation metrics}. For evaluation, we rely on quantitative metrics, like the $F_1$ score for classification and \textit{Mean Squared Error (MSE)} for regression, to measure the predictive performance of the black-box models. To assess a surrogate model's approximation in a global explainability setting, we use \textit{Global Fidelity (GF) defined as $ \mathrm{GF} = \frac{1}{N} \sum_{i=1}^N ( g_{\boldsymbol{\phi}} (\boldsymbol{x}_i) - f_{\boldsymbol{\theta}} (\boldsymbol{x}_i) )^2$}. Furthermore, to measure how good a surrogate model is at approximating the black-box model in a local neighborhood $N_{\boldsymbol{x}}$ of a point $\boldsymbol{x}$, which usually consists of synthetically generated perturbations of $\boldsymbol{x}$'s feature values \cite{Plumb}, we use \textit{Neighborhood Fidelity}. We obtain a ``global'' measure of neighborhood fidelity for the entire dataset, denoted as $\mathrm{GNF}$, by averaging across all data points: $\mathrm{GNF} = \frac{1}{N} \sum_{i=1}^N \frac{1}{|N_{\boldsymbol{x}}|} \sum_{\boldsymbol{x}^\prime \in N_{\boldsymbol{x}}}[( g_{\boldsymbol{\phi}} (\boldsymbol{x}^{\prime}) - f_{\boldsymbol{\theta}} (\boldsymbol{x}^{\prime}) )^2]$.

\begin{figure}[t!]
    \centering
    \subfloat{\includegraphics[scale=0.235]{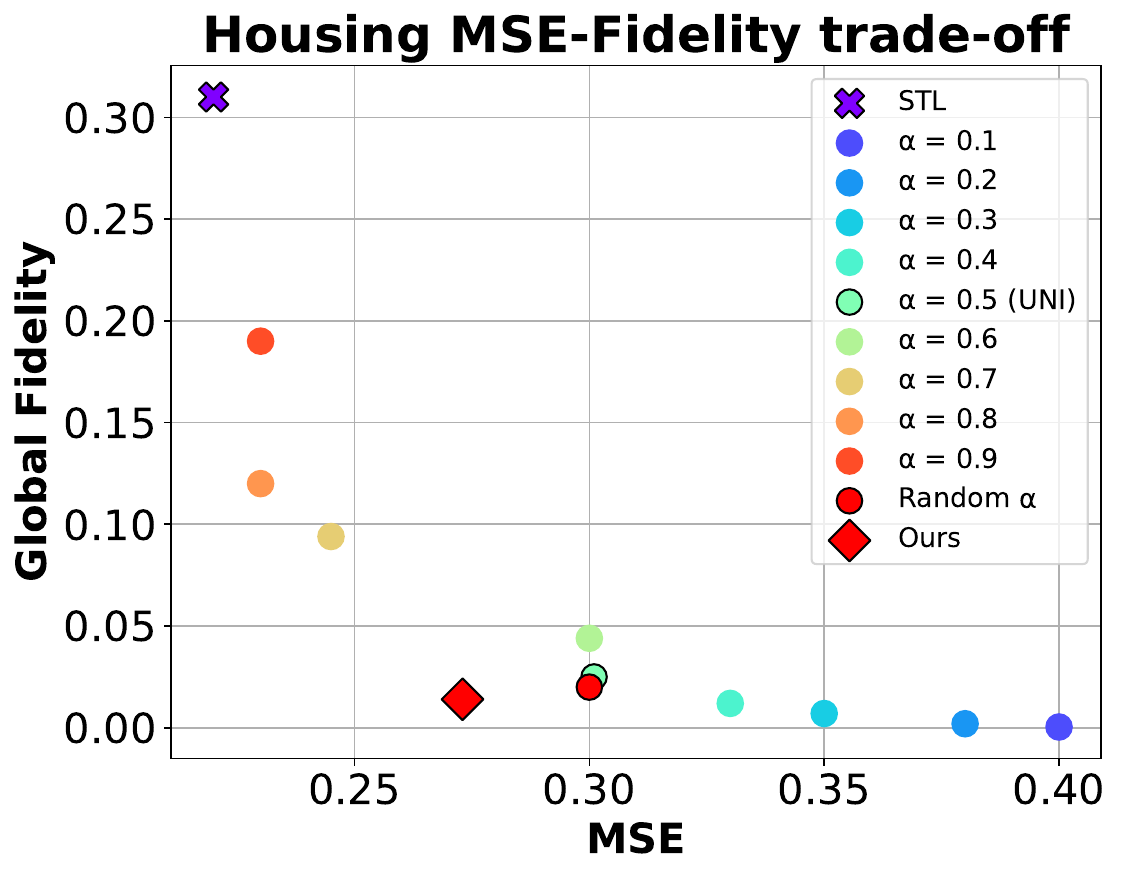}}\subfloat{\includegraphics[scale=0.235]{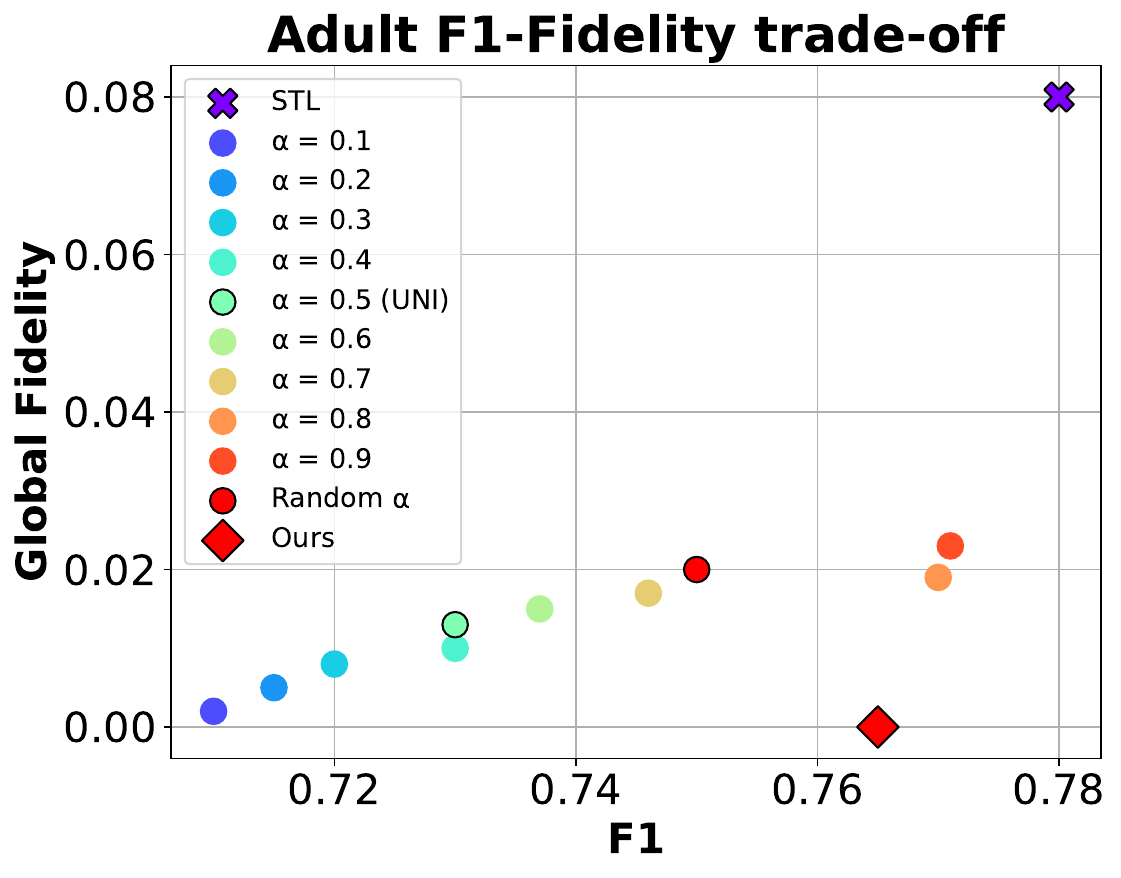}}\qquad
    \caption{Visualization of \textit{Predictive performance} vs. \textit{Global Fidelity} results for the \textsc{housing} (bottom-left is better) and the \textsc{adult} (bottom-right is better) datasets.}
    \label{fig:tradeoff} 
\end{figure}

\subsection{Global Explainability evaluation}
Global explainability provides insights for the overall behavior of the model across the entire dataset. It helps to identify which features the black-box model relies on the most. Table~\ref{table:baselines} shows the results of the experiments on the test set of each tabular dataset for our framework, as well as the baselines \textit{Linear}, \textit{STL}, \textit{UNI}, and \textit{RND}. The results showcase the superiority of our approach in terms of \textit{Global Fidelity} (lower is better), over all other baselines. For example, the improvement when compared to \textit{STL} is above $99\%$ for the \textsc{adult} dataset, while for the same dataset the (absolute) reduction in accuracy is less than $2\%$. Additionally, \textit{UNI} and \textit{RND} achieve worse performance in every metric compared to our approach. Finally, to justify the need for a ``black-box'' model, we also provide prediction test scores for a linear model, showcasing worse performance than the MLP model used in all experiments.

We also experiment with fixed values of the parameter $\alpha$ (\textit{grid search}).  Specifically, we experiment with different values of $\alpha$ in $(0, 1)$ with $step=0.1$, resulting in $9$ values in total, and we visualize the performance as a scatter plot of our evaluation metrics for two datasets in Fig.~\ref{fig:tradeoff}. Our method can find a local Pareto optimal solution, offering a ``best of both worlds'' scenario by finding points on the Pareto front with very low $\mathrm{GF}$ scores and minimal reductions in predictive performance, thus avoiding poor compromises in performance. In contrast, the \textit{grid search} method requires extreme values of $\alpha$ (e.g., $\alpha < 0.3)$ to achieve similar $\mathrm{GF}$ values, which results in a significant decline in the accuracy of the black-box model, demonstrating its inability to find Pareto-dominant solutions across the Pareto front. These results highlight the advantage of determining the value of $\alpha$ using our algorithm, despite the increased computational complexity introduced by the need to solve an optimization problem to find the optimal value $\alpha^*$.
\begin{table}[t!]
\caption{\label{table:local_baselines} Comparison of our approach with {\textit{STL}} and \textit{MTL} (\textit{UNI}, \textit{RND}) baselines with regards to $\mathrm{GNF}$.}

\centering
\adjustbox{width=\columnwidth}{
\hspace*{0.35cm}
\begin{tabular}{|c|cccc|c}
\cline{1-5}
\multicolumn{1}{|c|}{\bf Dataset} &
  \multicolumn{1}{c|}{\bf STL} &
  \multicolumn{3}{c|}{\bf MTL} &
   \\ \cline{1-5}
\multicolumn{1}{|c|}{} &
  \multicolumn{1}{c|}{} &
  \multicolumn{1}{c|}{\bf UNI } &
  \multicolumn{1}{c|}{\bf RND} &
  \multicolumn{1}{c|}{\bf Ours} 
   \\ \cline{1-5}
\multicolumn{1}{|c|} 
{\begin{tabular}[c]{@{}l@{}} \sc power  \\ \sc adult\\ \sc housing\\ \sc magic\\ \sc mnist \\ \sc mnist(cnn) \end{tabular}} & 
\multicolumn{1}{c|}{\begin{tabular}[c]{@{}c@{}} $2.1\cdot10^{-3}$ \\ $0.16$ \\ $5.67$\\ $0.12$ \\ $0.05$ \\ $0.07$ \end{tabular}} &
\multicolumn{1}{c|}{\begin{tabular}[c]{@{}c@{}} $1 \cdot10^{-3}$\\ $0.13$ \\ $0.18$\\ $0.10$ \\ $0.02$ \\ $0.005$ \end{tabular}} &
\multicolumn{1}{c|}{\begin{tabular}[c]{@{}c@{}} $2 \cdot10^{-3}$\\ $0.18$\\$ 0.11$ \\ $0.12$ \\ $0.05$ \\ $0.004$ \end{tabular}} &
\multicolumn{1}{c|}{\begin{tabular}[c]{@{}c@{}}$\bf5.7\cdot10^{-4}$ \\ $\bf0.038$ \\ $\bf0.085$\\ $\bf0.006$ \\ $\bf 0.003$ \\ $\bf 0.003$
\end{tabular}} &    \\ \cline{1-5}
\end{tabular}
}
\end{table}

\subsection{Local Explainability evaluation} 
Local explainability provides insights into how the black-box model makes predictions for individual instances, allowing users to interpret and trust specific decisions. To evaluate the approximation capability of our approach in a local explainability setting, we assess whether a surrogate model can more accurately approximate the black-box model $f_{\boldsymbol{\theta}^*}$ obtained by Algorithm~\ref{Alg:Alg1}, compared to black-box models obtained by the baseline algorithms we use in this paper, for each test instance separately. Specifically, we first train a black-box model $f_{\boldsymbol{\theta}^*}$ with our approach and the baseline algorithms, and then we select a local surrogate method (e.g., LIME\cite{Ribeiro} or SHAP \cite{Lundberg}) to approximate its output for each test instance. 
\par Additionally, our approach can be extended and can demonstrate improved results on more complex and domain-specific datasets, such as X-ray images. To this end, we also experiment, in the local explainability setting, with the \textsc{mnist} dataset. Specifically, we create a binary version of the image classification task (i.e., two labels) where the task is to predict whether an image depicts a specific digit or not, with one digit being the target class and the remaining ones being considered as the negative class. During training, each image is flattened to create an input vector wherein each feature is a pixel, and the goal of the surrogate model is to approximate the predictions of the black-box one.
\par We evaluate the approximation of the surrogate model using the $\mathrm{GNF}$ metric. Concretely, to calculate $\mathrm{GNF}$ for a data point $\boldsymbol{x}$, we generate $10$ neighbors ($|N_{\boldsymbol{x}}|=10$) by adding random Gaussian noise $\epsilon \sim \mathcal{N}(\boldsymbol{x}, \mu=0, \sigma^2=0.1)$ for tabular data and using random image patch deletion for \textsc{mnist}, following perturbation-based approaches \cite{Ribeiro, Lundberg}. For the local explainability method, we opt for LIME \cite{Ribeiro}. In Table~\ref{table:local_baselines}, we present the results of the experiments on local explainability for tabular data and images. The results for the predictive performance of $f_{\boldsymbol{\theta}^*}$ ($F_1, \mathrm{MSE}$) are omitted as they are the same with the ones in Table~\ref{table:baselines}, since we use the same training procedure for the black-box models.
\par The results demonstrate the improvement of our approach in the local surrogate model's fidelity compared to all baseline methods, on all datasets, as measured by the $\mathrm{GNF}$ metric. For instance, we notice improvements of up to $98.5\%$ in the \textsc{housing} dataset. 
Finally, in experiments with the \textsc{mnist} dataset, we observe improvements in $\mathrm{GNF}$ of up to $94\%$ compared to values obtained when black-box models trained by baseline algorithms were used as a base for the local surrogate. This means that a local surrogate model can better approximate a black-box model when trained using our algorithm. 

\subsection{Assessing the accuracy - approximability trade-off}
We conducted an ablation study to investigate two alternative strategies for optimizing the trade-off between approximability and accuracy. Our goal is to highlight the benefits of our proposed joint training approach which leverages MOO to effectively balance these competing objectives.
For the first strategy (\textit{J-SEP}), we jointly trained the two models and update their parameters only with their respective loss functions, i.e., the black-box model with the predictive loss and the surrogate one with Point Fidelity. In this experiment the approximation is negatively affected, indicating that the joint objective is crucial when updating the parameters of the black-box model to achieve better approximation by the surrogate one.

For the second strategy (\textit{J-DIST}), we utilized a form of knowledge distillation which, in simple terms, works by transferring the knowledge from a well-trained model (often called the ``teacher'') to another model (the ``student''). In our case, we trained a black-box model in the typical single-objective scenario to find an optimal solution $\boldsymbol{\theta}^*$ for the prediction task. Then a copy of this model $\boldsymbol{\theta}^\prime$ was made, and it was concurrently trained with the surrogate $\boldsymbol{\phi}$ using the joint convex objective we introduced, consisting of $\mathcal{L}_{\text{\tiny pred}}(\cdot)$ and $\mathcal{L}_{\text{\tiny PF}}(\cdot, \cdot)$, with the addition of the extra term $\mathcal{L}_{\text{\tiny dist}}(\boldsymbol{\theta}^*, \boldsymbol{\theta}^\prime) = \mathcal{L}_{\text{\tiny dist}}(f_{\boldsymbol{\theta}^*}(\boldsymbol{x}), f_{\boldsymbol{\theta}^\prime}(\boldsymbol{x})) = \left( f_{\boldsymbol{\theta}^*}(\boldsymbol{x}) - f_{\boldsymbol{\theta}^\prime}(\boldsymbol{x}) \right)^2$ which measures the distance between the predictions of the two black-box models (i.e., the “teacher” and the “student”):  $\mathcal{L} = \frac{1}{2} [ \mathcal{L}_{\text{\tiny pred}} (\boldsymbol{\theta}^\prime) + \mathcal{L}_{\text{\tiny dist}}(\boldsymbol{\theta}^*, \boldsymbol{\theta}^\prime)]+ \frac{1}{2}\mathcal{L}_{\text{\tiny PF}}(\boldsymbol{\theta}^\prime, \boldsymbol{\phi})$.
The purpose of this approach was to keep the predictions of the new model ($f_{\boldsymbol{\theta}^\prime}$) as close as possible to those of the optimal one ($f_{\boldsymbol{\theta}^*}$).

The inferior performance of both \textit{J-DIST} and \textit{J-SEP} approaches across predictive and approximability metrics in Table~\ref{table:ablations} underscores the necessity of our approach. Intuitively, this means that the joint objective manages to ``confine'' the black-box model's parameters, so that they can be better approximated by the surrogate one, without largely compromising its predictive performance.

\begin{table}[t!]
\caption{\label{table:ablations}  Comparison of knowledge distillation (\textit{J-DIST}) and sequential training (\textit{J-SEP}) with our approach, using the task-specific metrics and $\mathrm{GF}$.}

\centering

\adjustbox{width=\columnwidth}{

\renewcommand{\arraystretch}{1.27}

\begin{tabular}{|l|cccc}
\cline{1-4}
\multicolumn{1}{|c|}{\bf Dataset} &
  \multicolumn{3}{c|}{\bf Method} &
   \\ \cline{1-4}
\multicolumn{1}{|c|}{} &
  \multicolumn{1}{c|}{\bf J-SEP} &
  \multicolumn{1}{c|}{\bf J-DIST} &
  \multicolumn{1}{c|}{\bf Ours}
   \\ \cline{1-4}
\multicolumn{1}{|c|}
{\begin{tabular}[c]{@{}l@{}} \sc power (mse)  \\ \sc adult ($F_1$)\\ \sc housing (mse)\\ \sc magic ($F_1$)\\\end{tabular}} & 
\multicolumn{1}{c|}{\begin{tabular}[c]{@{}c@{}}$0.05$ ($\pm0.004$) \\ $0.78$ ($\pm0.02$) \\$0.23$ ($\pm0.02$)\\ $0.86$ ($\pm0.002$)\end{tabular}} &
\multicolumn{1}{c|}{\begin{tabular}[c]{@{}c@{}} $0.05$ ($\pm0.004$) \\ $0.78$ ($\pm0.02$) \\ $0.23$ ($\pm0.02$) \\ $0.86$ ($\pm0.002$) \end{tabular}} &
\multicolumn{1}{c|}{\begin{tabular}[c]{@{}c@{}}$0.06$ ($\pm0.002$)\\ $0.77$ ($\pm0.001$)\\ $0.27$ ($\pm0.003$) \\ $ 0.82$ ($\pm0.001$)\end{tabular}} &
\\ \cline{1-4}
\multicolumn{1}{|c|} 
{\begin{tabular}[c]{@{}l@{}} \sc power ($\mathrm{GF}$) \\ \sc adult ($\mathrm{GF}$)\\ \sc housing ($\mathrm{GF}$) \\ \sc magic ($\mathrm{GF}$) \\\end{tabular}} & 
\multicolumn{1}{c|}{\begin{tabular}[c]{@{}c@{}}$0.05$ ($\pm0.01$)\\ $0.09$ ($\pm0.0003$) \\$0.31$ ($\pm0.01$)\\ $0.05$ ($\pm0.002$)\end{tabular}} &
\multicolumn{1}{c|}{\begin{tabular}[c]{@{}c@{}} $0.005$ ($\pm0.0001$)\\ $0.01$ ($\pm0.0005$) \\ $0.08$ ($\pm0.002$) \\ $0.04$ ($\pm0.001$) \end{tabular}} &
\multicolumn{1}{c|}{\begin{tabular}[c]{@{}c@{}} $\bf8.91\cdot10^{-4}$ ($\bf\pm0.0015$) \\  $\bf1.2\cdot10^{-5}$ ($\bf\pm0.004$) \\ $\bf0.014$ ($\bf\pm0.02$) \\ $\bf3\cdot10^{-5}$ ($\bf\pm0.001$)\end{tabular}} &
\\ \cline{1-4}
\end{tabular}
 }
\end{table}

\subsection{Takeaways}

The key takeaways from our experiments can be summarised as follows:
\begin{itemize} 
\item The \textit{STL}-based baseline exhibits the worst approximation capability among all baselines, since it treats each objective separately. This limitation renders it inadequate for domains where explainability is crucial.
\item The \textit{MTL}-based approaches achieve the worst predictive accuracy among all baselines, rendering them unsuitable for domains where achieving high accuracy is necessary.
\item In all tasks, our MOO-based approach generates a local Pareto optimal solution for both objectives that achieves the best performance in \textit{Fidelity}, with improvements exceeding $99\%$ compared to the single-task baseline.
\item Our approach can be utilized also in the setting of local explainability, providing better approximation capabilities for both tabular data and images.

\end{itemize}
\section{Conclusion and Future Work}
\label{Conclusion}

We introduced a novel perspective on explainability, by modeling the trade-off between their approximation capability and the predictive performance of the black-box model as a MOO problem. We derived a joint training scheme wherein the parameters of the black-box model are jointly optimized via a gradient-based algorithm for the two objectives, so that it can be better approximated by a simpler interpretable surrogate model. Our results showed over $99\%$ improvement in \textit{Global Fidelity} and $98.5\%$ in \textit{Global Neighborhood Fidelity}.
\par Future research could evaluate the quality of explanations provided by the surrogate model beyond fidelity-based metrics. Furthermore, a compelling extension of this work would be its adaptation to the Federated Learning (FL) setting where the data are distributed and private. Adapting our MOO algorithm to the FL setting involves several challenges, notably non-independent and identically distributed data, which complicates the optimization process, as it requires to balance the two conflicting objectives of accuracy and explainability across clients with heterogeneous objectives.

\begin{Acknowledgments}
\par This work was conducted in the context of the Horizon Europe project PRE-ACT (Prediction of Radiotherapy side effects using explainable AI for patient communication and treatment modification). It was supported by the European Commission through the Horizon Europe Program (Grant Agreement number 101057746), by the Swiss State Secretariat for Education, Research and Innovation (SERI) under contract number 22 00058, and by the UK government (Innovate UK application number 10061955).
\end{Acknowledgments}

\end{document}